\documentclass{article}

\usepackage{arxiv}

\usepackage[utf8]{inputenc} 
\usepackage[T1]{fontenc}    
\usepackage{hyperref}       
\usepackage{url}            
\usepackage{booktabs}       
\usepackage{amsfonts}       
\usepackage{nicefrac}       
\usepackage{microtype}      
\usepackage{lipsum}
\usepackage{multirow}
\usepackage{graphicx}
\graphicspath{ {./images/} }

\title{Zero Touch Predictive Orchestration: Automating Time-Series Models for the Cloud-Edge Continuum}

\author{
 Abd Elghani Meliani \\
  Eurecom\\
  Campus SophiaTech, Biot, France \\
  \texttt{meliani@eurecom.fr} \\
   \And
 Arora Sagar \\
  OpenAirInterface\\
  Campus SophiaTech, Biot, France \\
  \texttt{sagar.arora@openairinterface.org} \\
  \And
 Adlen Ksentini \\
  Eurecom\\
  Campus SophiaTech, Biot, France \\
  \texttt{ksentini@eurecom.fr} \\
   \And
  Raymond Knopp \\
  Eurecom\\
  Campus SophiaTech, Biot, France \\
  \texttt{Knopp@eurecom.fr} \\ 
}

\begin{document}
\maketitle
\begin{abstract}
 The Cloud-Edge Continuum (CEC) enables latency-critical applications by distributing resources to the far edge, but its extreme volatility makes proactive Zero Touch Management via time-series forecasting essential. However, orchestrators face a severe "cold start" problem: newly discovered nodes lack the historical data required to train localized predictive models, while generalized models fail to capture unique hardware and microservice behaviors. To solve this, we propose a fully automated time-series prediction architecture driven by a novel data-mixing methodology. At the infrastructure level, we introduce a lightweight, technology-agnostic Resource Exposer (RE) that dynamically discovers nodes and continuously collects customizable telemetry (e.g., compute, network, energy). To overcome the sparsity of these initial local samples, our framework automatically merges them with TimeTrack \cite{timetrack}—our publicly available, high-resolution dataset collected at 45-second intervals. This synergizes TimeTrack’s foundational, high-frequency temporal patterns with the precise calibration of the local node data. Processed through a Neural Architecture Search (NAS) engine, the system automatically generates highly accurate baseline models. Experimental results demonstrate that merging the target data with TimeTrack effectively mitigates the cold start challenge. This specific integration significantly improves forecasting accuracy—measured in Mean Squared Error (MSE), Mean Absolute Error (MAE), and Mean Absolute Percentage Error (MAPE)—and accelerates convergence compared to training on the sparse local samples alone, training solely on generic datasets, or mixing the target data with standard alternative datasets, establishing a robust foundation for continuous MLOps deployment.
\end{abstract}

\section{Introduction}
\label{sec:introduction}
Traditional centralized cloud computing has driven digital transformation for over a decade, offering virtually limitless computational power. However, as new classes of applications emerge—such as augmented and virtual reality, autonomous vehicles, Federated Learning, and real-time video analytics—the centralized model is increasingly bottlenecked.
These modern workloads are highly sensitive to high latency, bandwidth limitations, and strict data-locality requirements, making a purely central deployment unfeasible. To address these limitations, the computing paradigm is rapidly shifting toward the Cloud-Edge Continuum (CEC) \cite{CECC}.
The CEC unlocks unprecedented capabilities. It allows compute-intensive tasks to benefit from the central cloud's abundance, while enabling latency-critical services to execute locally at the network edge with near-zero latency.  While the CEC provides immense architectural flexibility, it also introduces profound orchestration challenges. Unlike traditional cloud environments where resources are relatively static and managed centrally, the CEC is characterized by extreme volatility and heterogeneity.
Edge servers and Far-edge devices frequently fluctuate in availability, joining or leaving the network within seconds due to mobility or battery constraints.

In such dynamic environments, traditional reactive resource management fails. Waiting for a node to exhaust its capacity before scaling or migrating services inevitably degrades application performance and breaches Service Level Agreements (SLAs). Consequently, infrastructure providers must adopt proactive Zero Touch Management (ZTM) \cite{migration} techniques—such as predictive autoscaling and preemptive service migration \cite{bagaa2026layer} —to anticipate impending bottlenecks or operational anomalies, acting before degradation occurs. Implementing this proactive strategy inherently relies on machine learning, specifically time-series forecasting models (such as LSTMs, RNNs, and GRUs). However, these models exhibit a strict dependency on high-quality data. Because usage patterns, hardware characteristics, and microservice behaviors vary drastically across the heterogeneous CEC, a model trained solely on generic, pre-existing datasets cannot accurately predict performance on unseen infrastructure. Furthermore, manually collecting data and training a bespoke model for every unique machine type is unscalable and computationally prohibitive. To manage this distributed environment efficiently, the entire machine learning pipeline must be fully automated. To overcome this barrier, we propose an architecture that relies on lightweight, plugin-based Resource Exposer (RE) \cite{exposer} at the local infrastructure level. This module dynamically discovers volatile local nodes and continuously collects small, real-time telemetry samples. Crucially, the mechanism is highly customizable; instead of being limited to standard compute metrics like CPU or memory utilization, it can capture any measurable localized indicator, such as node-level energy consumption, carbon footprint, thermal thresholds, and network latency. Ultimately, the RE functions as an automated, multi-purpose data pipeline for the target machine.

While training a forecasting model exclusively on data from the specific target machine yields optimal performance, time-series models inherently require a substantial volume of historical data to achieve acceptable accuracy. Collecting a sufficient number of samples directly from a newly discovered source node requires a lengthy observation period, introducing an operational delay that contradicts the rapid, dynamic requirements of the CEC. To bypass this data-collection latency while still striving for localized accuracy, our architecture employs a strategic data-mixing approach. The system automatically merges the sparse, real-time target data collected by the Exposer with TimeTrack \cite{timetrack}, a foundational time-series dataset we collected from an operational OpenAirInterface (OAI) \cite{openairinterface} Kubernetes cluster used generally for running CI/CD workloads \cite{openairinterface_test}. This combination is highly synergistic: TimeTrack provides the algorithms with the foundational structural knowledge—teaching the model the complex temporal mechanics of when a metric is likely to surge or drop—while the small batch of specific data from the target machine calibrates the model to predict the exact magnitudes and values unique to that local environment. While other public monitoring datasets exist, they are generally not intended for time-series model training due to coarse, five-minute collection intervals. In contrast, TimeTrack is a high-resolution dataset purpose-built for training, offering a wide variety of fine-grained compute, network, and operational metrics at 45-second intervals. Supported by the Shannon–Nyquist sampling theorem \cite{Shannon}, this high frequency ensures the capture of transient behaviors, sudden bursts, and short-lived spikes that standard datasets inevitably smooth over.

Because many developers treat AI as a black box, the prevailing approach for training Deep Neural Networks (DNNs)—including time-series models—remains a manual, trial-and-error process of testing multiple architectures to select the highest-performing design. To fully automate this tedious selection, our framework integrates Neural Architecture Search (NAS) \cite{nas} methods. NAS is an automated machine learning technique that algorithmically explores a predefined search space to engineer custom network topologies for a specific dataset. By evaluating candidate configurations automatically, it eliminates the need for manual architectural engineering. Feeding our mixed, high-resolution dataset into this NAS engine completely automates initial model generation. Experimental results demonstrate that models initialized through this pipeline achieve significantly better accuracy and converge much faster than standard baselines, providing a robust, deployment-ready foundation for continuous MLOps fine-tuning.

In summary, the core contributions of this work are as follows:

\begin{itemize}
    \item \textbf{An Automated Prediction Architecture and Validation:} We propose an end-to-end ZTM framework that automates custom time-series model generation for diverse forecasting targets ranging from resource utilization to energy consumption. Experimental benchmarking demonstrates that models generated via our framework achieve superior accuracy—measured in Mean Squared Error (MSE), Mean Absolute Error (MAE), and Mean Absolute Percentage Error (MAPE)—and faster convergence compared to traditional baselines.
    \item \textbf{An Agnostic Resource Exposer (RE):} A plugin-based module deployed within local infrastructures. It allows volatile nodes to be dynamically discovered and exposes highly customized telemetry (e.g., compute, network, or energy metrics) to fuel the automated predictive modeling pipeline.
    \item \textbf{The TimeTrack Dataset:} A comprehensive, high-resolution time-series dataset featuring continuous computing and network monitoring at 45-second intervals, capturing the short-term dynamics necessary to teach structural forecasting patterns. The dataset is publicly available on Kaggle and Zenodo \cite{kagle, zenodo}, accumulating approximately 700 downloads in less than a year across both platforms.
\end{itemize}

The remainder of this paper is structured as follows. Section \ref{motivation} presents the motivation and background, using an edge deployment scenario to demonstrate the problem and justify the need for telemetry exposure and high-resolution baselines. Section \ref{sota} reviews related work across three domains—predictive orchestration, resource monitoring, and datasets—to contextualize our framework. Section \ref{methodology} details the architecture, encompassing the Resource Exposer, TimeTrack dataset, and NAS-driven model generation pipeline. Section \ref{evaluation} presents a two-phase evaluation of the data acquisition layer's footprint and the automated ML pipeline's performance. Finally, Section \ref{conclusion} concludes the paper and outlines future research directions.
\section{Motivation and Background}
\label{motivation}

To better understand the necessity of our fully automated architecture, it is essential to first examine the practical challenges of deploying predictive models in the CEC. In this section, we present a real-world scenario from our previous work that highlights the severe limitations of relying on standard, coarse-grained datasets for proactive resource management, and demonstrates why dynamic data exposure and high-resolution baselines are strictly required.

\subsection{The Generalization Challenge in Proactive Orchestration}
\begin{figure}[htbp]
    \centering
    \includegraphics[width=0.5\columnwidth]{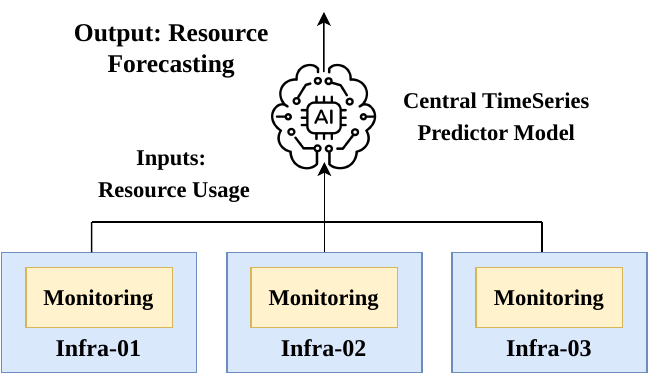}
    \caption{High-level architecture of the proactive resource forecasting system presented in \cite{migration}.}
    \label{fig:datacenter}
\end{figure}

In our previous work \cite{migration}, we developed a proactive system to manage the lifecycle of stateful microservices. The core idea was straightforward but effective: when a machine could no longer provide sufficient resources to meet its Service Level Agreements (SLAs), the system needed to preemptively relocate the microservice to another host. To achieve this, we designed a resource forecasting model using an LSTM time-series architecture. The model took historical CPU and memory usage as input and predicted future resource demands. A high-level architecture of this approach is illustrated in Figure \ref{fig:datacenter}. For the initial training, we relied on the MaternaGWA-13 \cite{gridWorkloadsArchive} dataset, which yielded promising accuracy on both the training and validation sets.

However, when we deployed this solution in a more realistic Cloud-Edge Continuum environment, the results degraded significantly. The fundamental reason for this failure was data distribution shift: the resource usage patterns in the generic GWA Materna dataset differed drastically from those observed on the new, highly volatile target machines. Consequently, the trained models struggled to generalize. This creates a critical \textbf{``cold-start'' generalization problem}: how to deploy an accurate predictive model on a newly discovered edge node without waiting weeks to collect a massive local historical dataset.

To overcome this issue and deploy accurate models in highly heterogeneous CEC environments, three potential strategies can be considered:
\begin{enumerate}
    \item \textbf{Training a separate model from scratch for each machine type.} While theoretically feasible, this is practically impossible in edge environments. It requires repeating the entire pipeline of manual data collection, cleaning, and training for every single node, resulting in unacceptable delays and massive computational overhead.
    \item \textbf{Fine-tuning the central model with new machine data.} This is more efficient than retraining from scratch but still requires manually collecting and preparing large amounts of target-specific data before the model becomes reliable enough for deployment.
    \item \textbf{Mixing data to overcome the cold-start phase.} Training a baseline model on a high-quality general dataset, augmented with a small, real-time sample of data collected directly from the target machine.
\end{enumerate}

The third solution is the only viable approach for the dynamic nature of the CEC. However, executing this strategy automatically requires solving two distinct challenges: i) how to retrieve the local target data dynamically, and ii) how to ensure the general baseline dataset is rich enough to teach structural patterns.

\subsection{The Need for Automated Local Exposure and High-Resolution Baselines}
To fulfill the first requirement—dynamically gathering the "small portion of samples" from the target machine—the system requires a dedicated, automated collection mechanism. This directly motivates the development of our \textbf{Resource Exposer (RE)}. Rather than relying on manual instrumentation, the RE acts as an agnostic plugin that automatically discovers new nodes and streams specific local telemetry (compute, network, or energy) back to the central pipeline.

\begin{figure}[htbp]
    \centering
    \includegraphics[width=0.5\columnwidth]{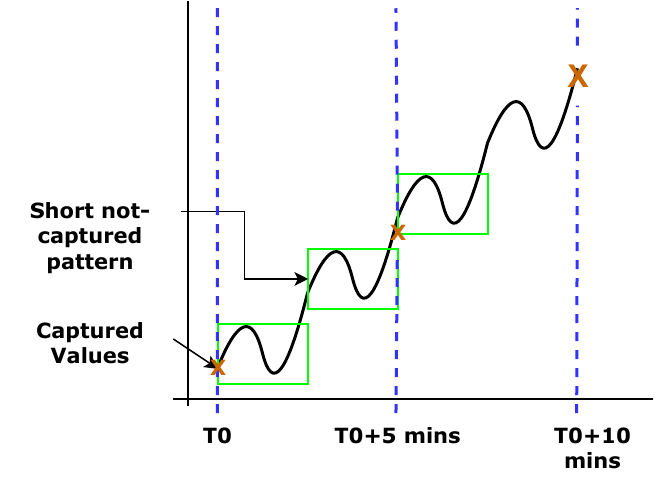}
    \caption{Example of critical short-term patterns that are smoothed over and lost due to coarse 5-minute data collection intervals.}
    \label{fig:patterns}
\end{figure}

However, mixing this local data with a public dataset is only effective if the public dataset accurately represents the rapid fluctuations of modern cloud-native systems. Unfortunately, most public datasets \cite{gridWorkloadsArchive, googleClusterData2011, azure_public_dataset_v1, alibabaClusterTrace2018, azure_public_dataset_v2} record metrics at coarse intervals, typically every 5 minutes. As illustrated in Figure \ref{fig:patterns}, significant variations—such as microservice scaling events or CPU bursts—occur within these intervals and are completely missed by coarse sampling.

From a signal processing perspective, the Shannon–Nyquist sampling theorem \cite{Shannon} states that to preserve the information carried by a signal, it must be sampled at least twice its highest frequency component. Sampling cluster dynamics every 5 minutes corresponds to sampling far below the effective frequency of edge operations, resulting in structurally impoverished sequences that limit the ability of models to learn meaningful temporal dependencies \cite{chuah2006sampling, libri2019towards, garland2019effects}. Guided by these principles, we collected \textbf{TimeTrack} at a 45-second interval. This resolution provides a practical balance: it is dense enough to capture transient behaviors and supply informative sequences to time-series models, yet lightweight enough to avoid monitoring overhead.

\subsection{The Proposed Automated Architecture}
\begin{figure}[htbp]
    \centering
    \includegraphics[width=0.7\columnwidth]{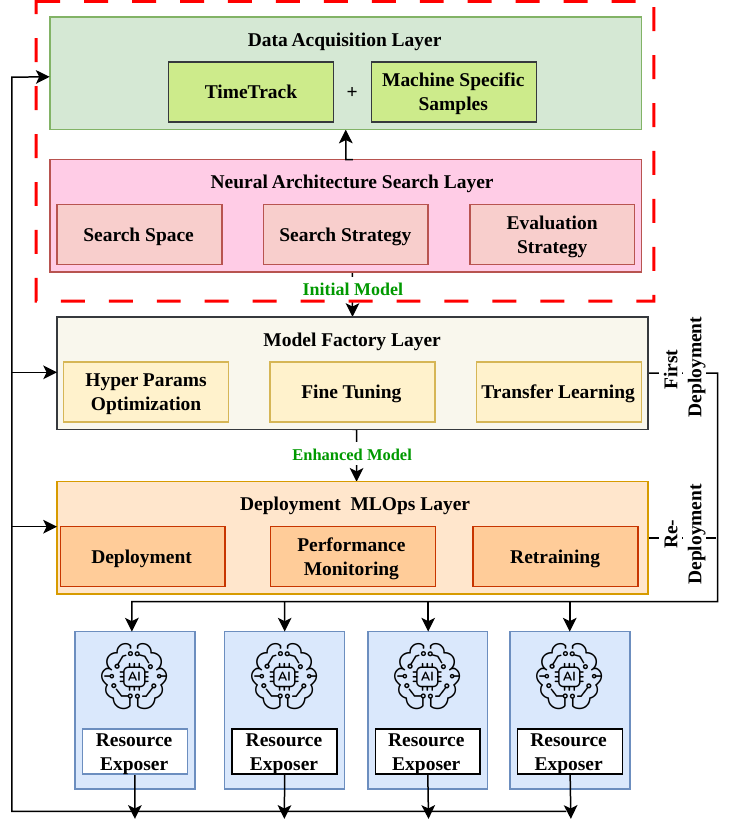}
    \caption{Overview of the proposed solution architecture for CEC environments. The first two layers (Data Acquisition via the Exposer, and NAS model generation) automatically combine local data with TimeTrack and are the primary focus of this work.}
    \label{fig:solutaion}
\end{figure}

By combining the automated local data collection of the Resource Exposer with the high-resolution structural baseline of TimeTrack, we establish a fully automated forecasting system. To clarify the role of these contributions within the broader vision of Zero Touch Management, Figure \ref{fig:solutaion} illustrates the end-to-end architecture we have built.

The first layer (Data Acquisition) utilizes the Resource Exposer to automatically collect samples from the target infrastructure and merge them with TimeTrack. This combined dataset is then passed to the second layer, where a Neural Architecture Search (NAS) generates an optimized initial model architecture from a predefined search space. Our experimental results confirm that this automated synergy yields models that are significantly more accurate and faster to deploy compared to existing methodologies.

\section{Related Work}
\label{sota}
To fully contextualize the contributions of our proposed ZTM framework, we categorize the related literature into three distinct domains:
i) automated predictive orchestration, ii) distributed resource monitoring, and iii) telemetry datasets.
\subsection{Automated Predictive Orchestration and MLOps}
The automation of resource management in cloud and telecommunications environments has
increasingly shifted from reactive, threshold-based heuristics to machine
learning-driven proactive methods. Numerous studies have explored the use of Deep
Learning architectures \cite{deeplearning}, such as LSTMs \cite{lstm} and GRUs, to forecast resource usage and trigger
proactive autoscaling or migration . However, the majority of these solutions assume a
static environment where a model, once trained on a historical dataset, maintains its accuracy over time. 

More recent works have recognized the need for continuous adaptation,
leading to the integration of Machine Learning Operations (MLOps) \cite{alla2020mlops} pipelines in 5G and edge environments.
These frameworks facilitate continuous learning by retraining models as new data arrives.
Furthermore, techniques such as AutoMl \cite{karmaker2021automl} Neural Architecture
Search (NAS) \cite{nas} and transfer learning have been utilized to automatically
discover optimal model structures and adapt them to new tasks.
Despite these advancements, existing automated orchestration frameworks fail
to address the "cold-start" generalization problem inherent to the CEC.
Current MLOps pipelines focus on managing models that are already deployed;
they do not automate the critical initialization phase—specifically,
how to intelligently mix high-resolution structural datasets with dynamic target
telemetry to generate a deployment-ready baseline model for unseen,
heterogeneous edge nodes. Our architecture bridges this exact gap.

\subsection{Resource Monitoring and Exposure Systems}
Building proactive models requires robust data collection mechanisms.
Numerous approaches have been proposed for systems monitoring in distributed
environments. One prominent example is the DECOR system \cite{decor},
which utilizes a distributed method for resource monitoring, particularly in
network-based applications like redundancy elimination and traffic sampling.
By distributing monitoring tasks across various network nodes, DECOR avoids the
bottlenecks typically associated with centralized controllers.
Similarly, Dprof \cite{dprof} presents a lightweight, distributed profiling system
designed to trace Remote Procedural Calls (RPC) operations and identify performance
bottlenecks. Dprof gathers and analyzes RPC traces from heterogeneous components,
storing them in a distributed data store. While effective for debugging at the RPC level,
Dprof focuses on system-level event profiling and lacks comprehensive infrastructure
monitoring capabilities. A scalable monitoring framework introduced in \cite{mekki}
specifically tackles the challenges of monitoring 5G network slices,
with an emphasis on resource isolation, multi-tenancy, and the integration
of different technological domains (e.g., RAN and cloud). Recently, both industry
and academia have been advocating for standardized resource exposure solutions to
improve infrastructure-aware service deployment, as emphasized by proposals from
the Internet Engineering Task Force (IETF) \cite{ietfmetrics}.
While we share the broader goal of improving resource management and service discovery,
existing tools are often deeply coupled with specific orchestrators.
We advance the state-of-the-art by introducing a unified, plugin-based Resource Exposer tailored for heterogeneous CEC systems, strictly separating resource exposure from service orchestration.
\subsection{Cloud and Edge Telemetry Datasets}
\begin{table*}[htbp]
    \centering
    \caption{Comparison of TimeTrack with Related Datasets, C:Compute, N:Network, S:Storage}
    \label{tab1}
    \begin{tabular}{|c|c|c|c|c|c|c|c|}
        \hline
        \textbf{Dataset} & \textbf{Nb Of machines} & \textbf{Collection Interval} & \textbf{Duration} & \textbf{Setup} & \textbf{Detail Level} & \textbf{Metrics} \\ \hline
        \cite{googleClusterData2011} & 12,500 & 5 min & 29 days & Virtual & medium & C \\ \hline
        \cite{alibabaClusterTrace2018} & 4,000 & 5 min & 8 days & Virtual & low & C  \\ \hline
        \cite{gridWorkloadsArchive} & 1,594 & 5 min & 3 months & Virtual & medium & C, N, S  \\ \hline
        \cite{azure_public_dataset_v1} & 2,013,767 & 5 min & 30 days& Virtual & low & C \\ \hline
        \cite{azure_public_dataset_v2} & 2,695,548 & 5 min & 30 days& Virtual  & medium & C \\ \hline
        \textbf{TimeTrack} & \textbf{7} & \textbf{45 sec} & \textbf{30 Days}  & \textbf{Physical} & \textbf{high}  & \textbf{C, N, S} \\ \hline
    \end{tabular}
\end{table*}

The foundational training of any automated forecasting architecture relies heavily on
the training data. Several publicly available datasets—such as Google Cluster Data 2011 \cite{googleClusterData2011},
Alibaba Cluster Traces \cite{alibabaClusterTrace2018}, Grid Workloads Archive GWA-T-13-Materna
\cite{gridWorkloadsArchive}, and Azure Public Traces \cite{azure_public_dataset_v1,
azure_public_dataset_v2}—provide valuable insights into resource utilization. 

However, as summarized in Table \ref{tab1}, these datasets possess limitations that reduce
their efficacy for predictive modeling in modern edge environments.
Many cover only short durations or focus exclusively on computing metrics, neglecting the
network and storage performance data essential for comprehensive analysis.
Furthermore, most related works rely on data collected in virtualized environments,
where virtualization overhead can obscure true hardware performance. 
Most critically, existing datasets rely on coarse data collection intervals—almost
universally 5 minutes. While sufficient for general historical monitoring,
this sparsity is inadequate for time-series forecasting in dynamic microservice
environments, as it fails to capture rapid fluctuations and transient system behaviors.
In contrast, our TimeTrack dataset was collected directly from physical machines at a
fine-grained 45-second interval. This high-resolution setup eliminates virtualization
artifacts and captures the exact, real-world transient behaviors necessary to pre-train
robust predictive models for the CEC.

\section{Automated Predictive Orchestration Architecture}
\label{methodology}
To address the ``cold-start'' generalization problem in heterogeneous CEC environments,
we propose a fully automated time-series forecasting pipeline. This framework intelligently
merges dynamic local telemetry with a high-resolution foundational dataset to automatically
generate optimized predictive models.
\subsection{End-to-End Pipeline Architecture}
\begin{figure}[htbp]
    \centering
    \includegraphics[width=0.8\columnwidth]{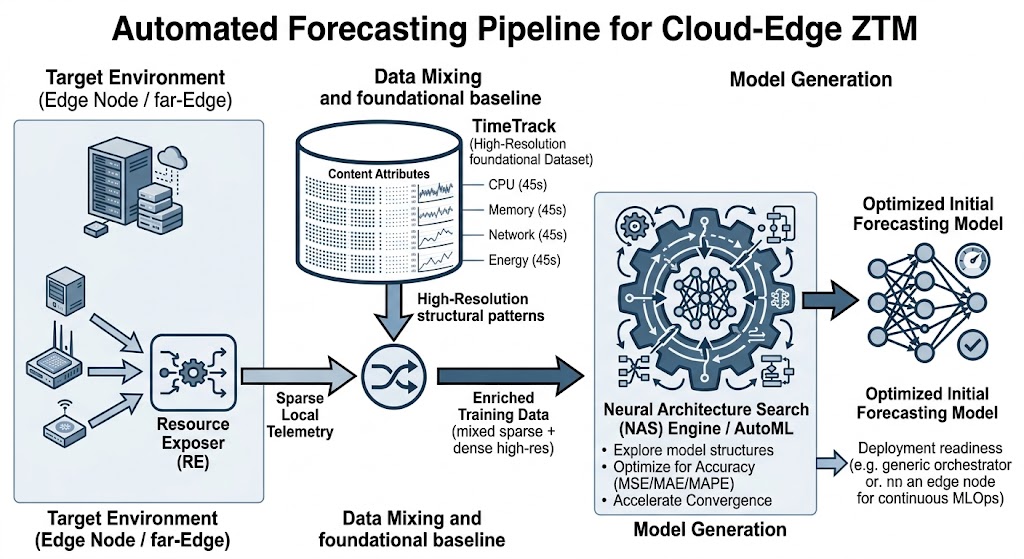}
    \caption{End-to-end architecture of the automated forecasting pipeline, illustrating the flow from System Discovery and the Resource Exposer, through Data Mixing with TimeTrack, to NAS Model Generation.}
    \label{fig:architecture}
\end{figure}

The core workflow of our proposed solution is illustrated in 
Figure \ref{fig:architecture}. The architecture is built upon three primary pillars
that operate sequentially to eliminate the need for manual model engineering:
\begin{enumerate}
    \item \textbf{System Discovery and Target Exposure:} The pipeline begins with a Discovery Phase,
     which automatically identifies newly available physical or virtual nodes and registers
    the specific metrics they are capable of providing (e.g., CPU, memory, energy type).
    Once discovered, the Resource Exposer (RE) extracts real-time, localized telemetry
    from the target environment.
    \item \textbf{Data Mixing and Foundational Baseline:} Because the initial telemetry
    gathered from a new target node is too sparse to train an accurate neural network
    from scratch, it is merged with \textbf{TimeTrack}, our pre-collected,
    high-resolution dataset. TimeTrack acts as the structural baseline,
    teaching the model the fundamental temporal mechanics of system fluctuations.
    \item \textbf{NAS-Driven Model Generation:} The enriched, mixed dataset is fed into
    an automated NAS engine. The NAS explores various time-series architectures to
    optimize for forecasting accuracy (MSE, MAE, MAPE), outputting a
    deployment-ready model tailored specifically to the newly discovered edge node.
\end{enumerate}

\subsection{Dynamic Data Acquisition: The Resource Exposer Framework}
\label{sec:exposer}

The dynamic data acquisition layer relies on our novel, lightweight Resource Exposure framework.
Deployed on target clusters, the RE is responsible for collecting real-time local
resource usage and exposing it via a northbound API to the central ML pipeline. 

\begin{figure}[htbp]
    \centering
    \includegraphics[width=0.8\columnwidth]{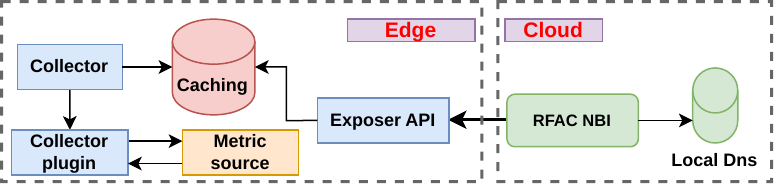}
    \caption{The lower-level architecture of the Resource Exposer, illustrating the plugin-based data collection and message broker integration.}
    \label{fig:low_level_arch}
\end{figure}

As illustrated in Figure \ref{fig:low_level_arch}, to handle the immense diversity of
monitoring solutions and infrastructure types in the CEC, the RE is built 
upon a \textit{gRPC plugin-based architecture}.
These plugins serve as intermediaries between the data source (e.g., Prometheus \cite{prometheus}, Nagios \cite{nagios}, Nvidia GPU exporters ... etc)
and the central collector. This design ensures that the RE is completely agnostic and
exceptionally customizable. It can be configured to expose any measurable target,
ranging from standard computing resources (CPU, GPU, memory, storage) and network data
(throughput, packet loss, latency) to infrastructure characteristics and energy
consumption metrics.

To ensure consistency across these diverse nodes, the RE employs a 
\textbf{unified data format}. Regardless of the underlying hardware or the specific
plugins used, the framework normalizes all metrics into a structured,
standard response. This uniformity is crucial for automated machine learning,
as it allows the NAS engine to ingest data from wildly different infrastructures
without requiring manual data-cleaning interventions.

The internal interaction workflow of the framework is designed for efficiency and responsiveness. As detailed in Figure \ref{fig:workflow}, the operational lifecycle consists of four key steps:

\begin{figure}[htbp]
    \centering
    \includegraphics[width=0.8\columnwidth]{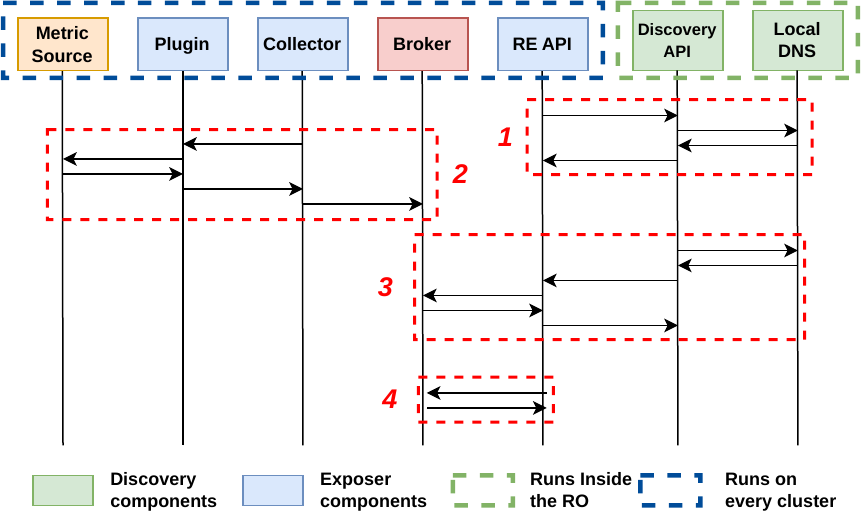}
    \caption{Interaction workflow between the different framework components, highlighting registration, continuous collection, exposure, and memory management.}
    \label{fig:workflow}
\end{figure}

\begin{enumerate}
    \item \textbf{Exposer Registration:} When a new RE is deployed, it uses a Local DNS system to self-register with the centralized Discovery Module. This decentralized method allows the system to scale flexibly, tracking nodes as they dynamically join or leave the network.
    \item \textbf{Data Collection:} The internal collector component periodically accesses source metrics via the plugins and temporarily stores them in a lightweight \textbf{message broker} (e.g., RabbitMQ, Kafka, Redis).
    \item \textbf{Data Exposure:} The Discovery Module requests specific information, prompting the RE to retrieve the latest metrics from the broker and expose them. This short-term caching mechanism ensures that the ML pipeline has instantaneous access to near real-time data under peak conditions.
    \item \textbf{Broker Purge:} To maintain optimal memory usage on constrained edge devices, the RE automatically purges the broker if the API remains unused for a defined period, preventing unnecessary data accumulation.
\end{enumerate}

\subsection{Foundational Baseline: The TimeTrack Dataset}
While the RE calibrates the model with real-time local constraints, the ML algorithms require a dense structural baseline to learn complex temporal dependencies. This is provided by \textbf{TimeTrack}, a high-resolution dataset collected over a 30-day period. We provide the following overview of its environment, structure, and key analytical patterns to assist researchers who may wish to utilize the dataset independently for their own machine learning and forecasting tasks. For a more exhaustive system-level analysis of the collected metrics, we refer the reader to the original dataset paper \cite{timetrack}.
\subsubsection{Environment and Data Structure}
TimeTrack was collected from an OpenAirInterface (OAI) testing cluster used for CI/CD pipelines of 5G components (e.g., gNB, UE, and CN). The physical cluster comprises seven machines equipped with a total of 437.5 GB of RAM, 236 CPU cores, 1800 GB of SSD storage, and 38 physical network interfaces. Utilizing the same RE plugin mechanism described above interfacing with Prometheus, data was collected at highly granular **45-second intervals**. The distribution of resources across the machines is detailed in Table \ref{tab:resource_dist}.

\begin{table}[htbp]
\centering
\caption{Resource Distribution Across the OAI Cluster Machines}
\begin{tabular}{|c|c|c|c|c|}
\hline
\textbf{Machine (No)} & \textbf{Cores} & \textbf{RAM (GB)} & \textbf{Disk (GB)} & \textbf{Physical IF (No)} \\
\hline
1 & 36 & 62.5 & 278.37 & 4 \\
2 & 48 & 62.5 & 222.50 & 6 \\
3 & 36 & 62.5 & 278.37 & 6 \\
4 & 36 & 62.5 & 278.37 & 4 \\
5 & 24 & 62.5 & 222.50 & 6 \\
6 & 36 & 62.5 & 278.37 & 6 \\
7 & 20 & 62.5 & 222.50 & 6 \\
\hline
\end{tabular}
\label{tab:resource_dist}
\end{table}

The dataset is available online \cite{zenodo}, \cite{kagle} consists of four comprehensive traces:
\begin{itemize}
    \item \textbf{Compute Metrics:} Available and used memory, average CPU consumption, and read/write disk throughputs at both cluster and machine levels.
    \item \textbf{CPU Core Utilization:} Granular availability and utilization tracking for each of the 236 individual CPU cores.
    \item \textbf{Network Latency Metrics:} Minimum, maximum, average, and mean deviation (mdev) of RTT, alongside jitter measurements.
    \item \textbf{Network Interface Metrics:} Dropped packets, error rates, and transmitted/received throughput for physical interfaces.
\end{itemize}

\subsubsection{Structural Analysis and ML Applicability}
\begin{figure*}[h]
    \centering
    \includegraphics[width=\textwidth]{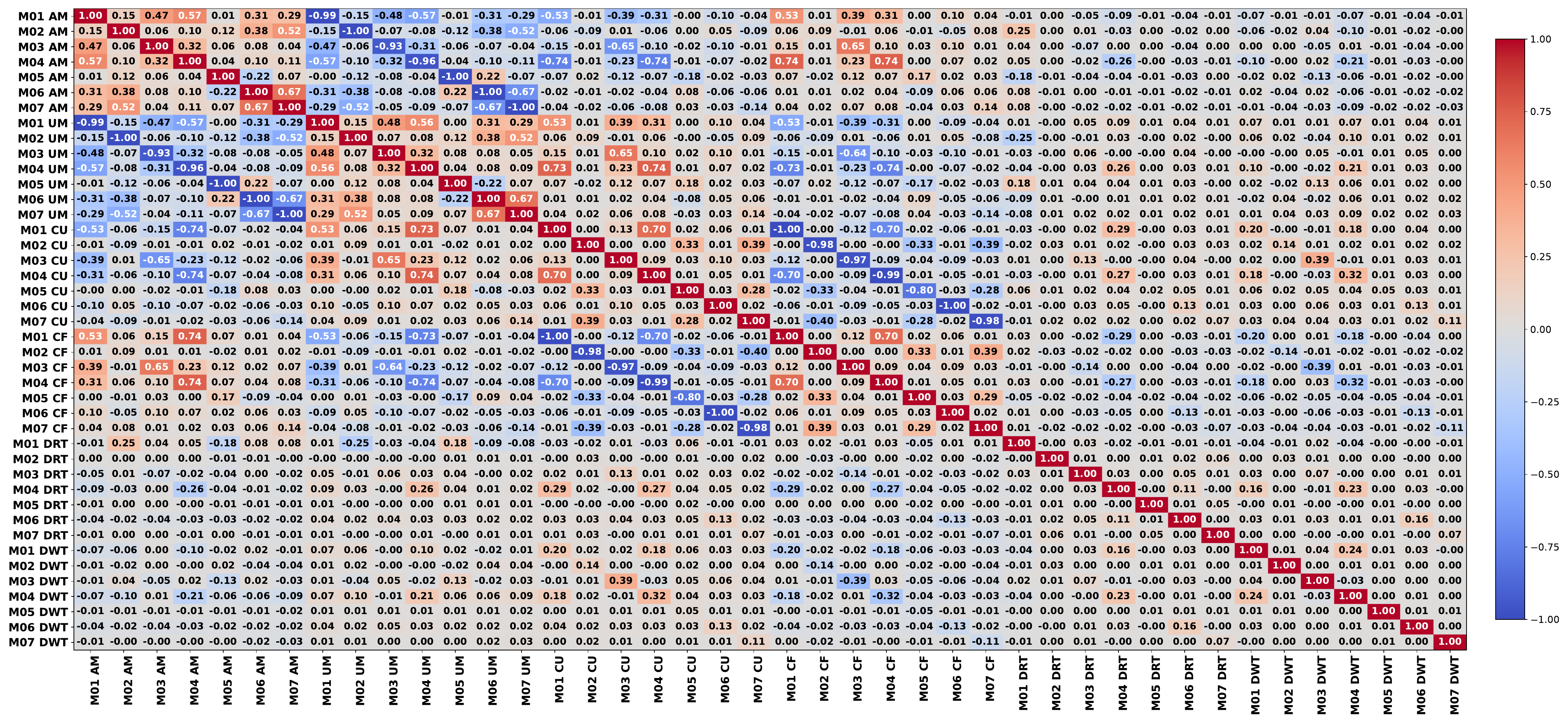}
    \caption{Correlation matrix demonstrating relationships between CPU usage (CU), memory usage (UM), disk throughput (DRT/DWT), and disk usage (UD).}
    \label{fig:correlation}
\end{figure*}

To validate TimeTrack's utility as a structural training baseline, we analyzed its temporal patterns and metric correlations. Figure \ref{fig:correlation} presents the correlation matrix for the compute metrics. 

The analysis reveals a notable positive correlation between CPU usage and memory consumption across several machines (e.g., $0.65$ for machine 03, and $0.74$ for machine 04). This indicates that the dataset successfully captures the coupled nature of CPU- and memory-intensive workloads, providing ML models with realistic multivariate relationships. 

\begin{figure}[htbp]
\centering
\includegraphics[width=0.5\columnwidth]{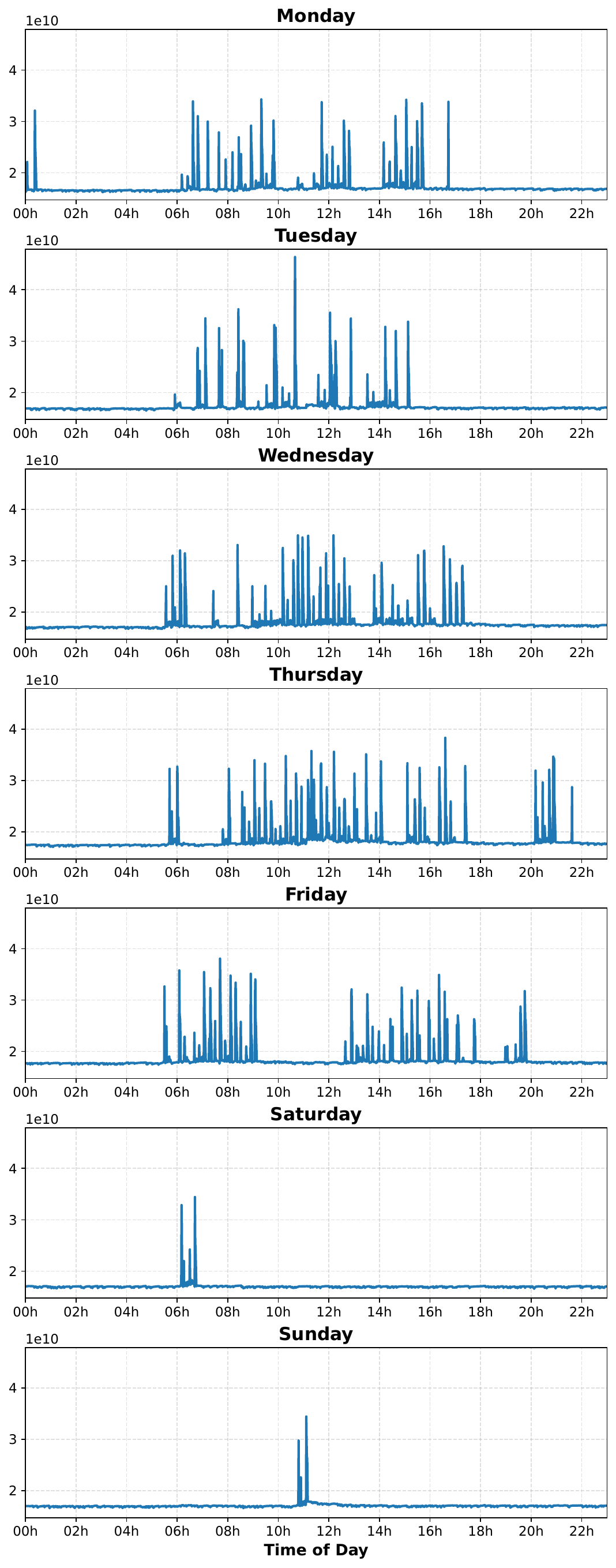}
\caption{Memory utilization across the cluster over a week, showing distinct usage spikes during weekdays versus low, stable background utilization over the weekend.}
\label{fig:weekly}
\end{figure}

Furthermore, forecasting models rely heavily on recognizing cyclical behavior. Figure \ref{fig:weekly} illustrates memory utilization across the cluster over a standard week. 

The temporal patterns are highly distinct: during the weekend, memory usage remains low and stable, reflecting an idle operational state. Conversely, during weekdays, utilization rises significantly during working hours, indicating high demand from active workloads, before returning to a stable baseline. This regular daily and weekly cycle provides highly predictable demand patterns. By feeding these cyclical patterns—along with the fine-grained 45-second transient behaviors—into our automated NAS pipeline, the resulting models inherit a robust structural understanding of infrastructure dynamics, allowing them to rapidly adapt to the new environments exposed by the RE.

\section{Evaluation and Results}
\label{evaluation}
To comprehensively assess the performance of our automated predictive orchestration 
framework, we divide our evaluation into two main phases. First, we profile the Resource
Exposer to ensure it operates efficiently on low-resource edge nodes without inducing
monitoring overhead. Second, we evaluate the end-to-end Machine Learning pipeline,
demonstrating how the integration of TimeTrack improves NAS efficiency, model accuracy,
and post-deployment computational constraints compared to widely used public datasets.

\subsection{Profiling the Resource Exposer on Edge Hardware}
\label{eval:exposer}
The RE must operate continuously on every computing node within the Cloud-Edge Continuum.
Therefore, it is strictly required to have a minimal resource footprint.
To validate its edge-readiness, we deployed the RE on a low-resource Raspberry Pi 4 model
(Broadcom BCM2711 quad-core Cortex-A72 64-bit CPU at 1.5 GHz, 8 GB RAM) running Ubuntu 22.04
and a standard containerd-based edge runtime. Redis Stack was utilized as the local message broker.

Because the RE is highly sensitive to the frequency of data extraction, we measured the CPU and memory consumption of both the exposer and the broker over a one-week period, varying the data collection interval from a standard 20 seconds down to an extreme 1-second interval. 

\begin{figure}[htbp]
    \centering
    \includegraphics[width=0.7\columnwidth]{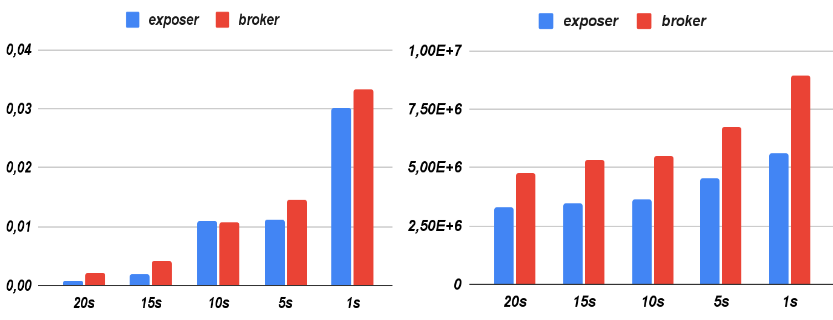}
    \caption{CPU consumption (left) and memory usage (right) of the Resource Exposer and the internal broker across varying data collection intervals.}
    \label{fig:exposer_broker_footprint}
\end{figure}

As illustrated in Figure \ref{fig:exposer_broker_footprint} (left), even under the most aggressive 1-second collection scenario, both the exposer and the broker consume less than 0.04 CPU cores. Figure \ref{fig:exposer_broker_footprint} (right) reveals that the exposer's memory footprint remains exceptionally stable regardless of the collection frequency. While the broker exhibits a slight increase in memory usage as the interval tightens to 1 second, it peaks at merely 8.51 MB. This extreme memory efficiency is a direct result of the broker purge mechanism detailed in Section \ref{sec:exposer}. 

\begin{figure}[htbp]
    \centering
    \includegraphics[width=0.5\columnwidth]{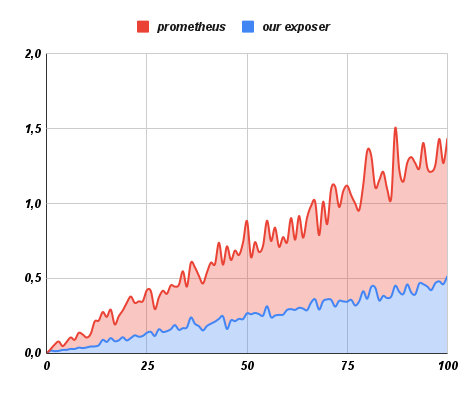}
    \caption{Comparison of API response times under concurrent request loads, demonstrating the efficiency of the RE's short-term caching versus Prometheus.}
    \label{fig:api_latency}
\end{figure}

Finally, we evaluated the response latency of the RE's northbound API. We benchmarked it against Prometheus \cite{prometheus}, a standard time-series monitoring toolkit, over 50 trials with varying concurrent request loads. As depicted in Figure \ref{fig:api_latency}, while response times for both systems increase under heavy concurrency, the Prometheus API takes approximately twice as long to respond. This validates our architectural choice to utilize a lightweight message broker for short-term caching, allowing the RE to serve near real-time telemetry significantly faster than solutions relying on heavy Time Series Databases (TSDBs).

\subsection{Automated CPU Model Generation and NAS Evaluation}
\label{eval:nas}

Having established the lightweight footprint of the data acquisition layer, we now evaluate the core automated model generation pipeline. To contextualize this evaluation, we leverage Neural Architecture Search, an automated subfield of machine learning designed to eliminate human bias and the tedious trial-and-error typically associated with manual network design. A NAS framework is fundamentally governed by two core components: the \textit{search space}, which defines the bounded universe of all valid network configurations, layers, and hyperparameters available for selection; and the \textit{search strategy}, which dictates the optimization algorithm used to navigate and sample from this space to discover high-performing architectures.

The primary objective of this subsection is to determine whether mixing a target node's sparse local data with a high-resolution structural baseline (TimeTrack, 45s interval) yields superior NAS convergence and predictive accuracy compared to standard, coarse-grained public datasets. For the purpose of this evaluation, we utilize CPU usage as our representative target metric. While CPU utilization frequently exhibits strong correlations with other system parameters, such as memory consumption as shown in figure \ref{fig:correlation}, we intentionally isolate CPU data to maintain a controlled experimental environment. The objective of this specific experiment is not to engineer an exhaustive, production-ready multivariate predictor, but rather to strictly isolate and compare the models' structural accuracy and convergence behavior across different underlying datasets. For comparative benchmarking, we selected the Grid Workloads Archive Materna-13 (GWA-M13) \cite{gridWorkloadsArchive} and the Alibaba Cluster Traces v2018 (ACTv2018) \cite{alibabaClusterTrace2018}, both of which rely on 5-minute sampling intervals.

To power the automated pipeline, we utilized Microsoft NNI \cite{nni2021} as the underlying NAS engine. The architectural search space (detailed in Table \ref{tab:searchspace}) was intentionally designed to be highly heterogeneous, encompassing Multi-Layer Perceptrons (MLP), standard Recurrent Neural Networks (RNN), Long Short-Term Memory (LSTM) networks, Gated Recurrent Units (GRU), Convolutional Neural Networks (CNN), Temporal Convolutional Networks (TCN), and Transformers. This diverse selection is justified by the complex, multi-scalar nature of cloud workload telemetry. While sequential models like LSTM and GRU excel at capturing long-term temporal dependencies and seasonal trends, CNNs and TCNs utilize parallelized receptive fields to isolate transient, short-term burstiness. Transformers leverage self-attention to map global contextual correlations across extended horizons, and MLPs provide a minimal non-linear baseline. By broadening the search space to include these fundamentally distinct inductive biases alongside micro-hyperparameters (e.g., window size, learning rate, and attention heads), we ensure that the pipeline can adaptively synthesize the optimal model topology for any given data composition.
\begin{table}[htbp]
\centering
\caption{Neural Architecture Search (NAS) hyperparameter search space.}
\label{tab:searchspace}
\begin{tabular}{lll}
\hline
\textbf{Parameter} & \textbf{Type} & \textbf{Values / Range} \\
\hline
\multirow{2}{*}{Model type} & \multirow{2}{*}{choice} & LSTM, GRU, CNN, RNN, \\
                            &                         & TCN, Transformer, MLP \\
Units & choice & 16, 32, 64, 128, 256 \\
Number of layers & choice & 1, 2, 3, 4 \\
Dropout & choice & 0.0, 0.1, 0.2, 0.3, 0.5 \\
Activation & choice & relu, tanh, sigmoid, elu \\
Learning rate & loguniform & $[10^{-5}, 10^{-2}]$ \\
Window size & choice & 2–30 \\
Batch size & choice & 16, 32, 64, 128 \\
Epochs & choice & 5, 10, 15, 20 \\
Kernel size & choice & 2, 3, 5, 7 \\
Attention heads & choice & 2, 4, 8 \\
\hline
\end{tabular}
\end{table}
To guarantee that our subsequent performance evaluations are algorithmically robust and invariant to the biases of a single optimization routine, we executed the experiments across five distinct search strategies, each representing a fundamentally unique paradigm of space exploration:
\begin{itemize}
    \item \textbf{Grid Search:} A deterministic, exhaustive strategy that systematically evaluates every predefined combination within the discrete bounds. It establishes a rigid performance baseline, though it is inherently constrained by the curse of dimensionality.
    \item \textbf{Random Search:} A non-deterministic strategy that samples configurations completely at random. Statistically, it serves as a powerful benchmark by exploring continuous dimensions (such as learning rate) more effectively than Grid Search without getting trapped in redundant parameter coordinates.
    \item \textbf{Tree-structured Parzen Estimator (TPE):} A Sequential Model-Based Optimization (SMBO) approach rooted in Bayesian inference. TPE constructs a probabilistic model of architecture performance based on historical evaluation history, actively guiding the search toward regions with a high probability of improvement while minimizing computational waste.
    \item \textbf{Evolutionary Algorithm:} A population-based heuristic inspired by natural selection. By maintaining a pool of architectures and applying iterative mutations and crossovers to the top-performing individuals, it is uniquely suited for navigating highly non-convex, discontinuous search spaces and escaping local minima.
    \item \textbf{Simulated Annealing:} A single-state metaheuristic inspired by metallurgy. It initially explores the search space aggressively by probabilistically accepting lower-performing candidate architectures (at a high virtual "temperature"), gradually cooling over time to tighten its focus around the global optimum.
\end{itemize}

By evaluating our proposed data-mixing methodology against this spectrum of exhaustive, stochastic, probabilistic, and heuristic search regimes, we ensure that the resulting predictive accuracies reflect genuine structural data advantages rather than artifacts of a specific optimization algorithm.

\subsubsection{Resource-Constrained NAS Performance and Accuracy}
\label{eval:resource_constrained}

Before deploying machine learning models in dynamic cloud environments, it is critical to ensure that the automated pipeline can bootstrap highly accurate initial models rapidly and without relying on expensive GPU clusters. To validate this, our first experiment evaluates the best outcome of all search methods within our defined search space using strictly limited computing resources (a standard commercial Intel Core i7 CPU) over severely restricted search intervals ranging from 5 to 30 minutes (300 to 1800 seconds). Demonstrating convergence within these short windows proves the framework's ability to provide robust initial models that can be immediately deployed and fine-tuned on the fly. 

To provide a holistic view of predictive performance, we evaluate the generated architectures using three distinct, complementary metrics:
\begin{itemize}
    \item \textbf{Mean Absolute Percentage Error (MAPE):} Our primary scale-independent metric, enabling uniform accuracy comparisons across target nodes with varying absolute resource capacities.
    \item \textbf{Mean Absolute Error (MAE):} Quantifies error in tangible operational units (e.g., absolute CPU percentages) to establish a concrete baseline of physical resource inaccuracy.
    \item \textbf{Mean Squared Error (MSE):} Disproportionately penalizes large deviations, explicitly highlighting a model's vulnerability to missing catastrophic workload spikes that risk Service Level Agreement (SLA) violations.
\end{itemize}

To conduct a rigorous ablation study and validate the necessity of our data-mixing approach, we executed the NAS search across seven distinct data scenarios. In the \textit{Mixed} scenarios, the NAS engine was constrained to the maximum of 1,500 samples: 1,000 structural samples from a generic dataset (TimeTrack, GWA-M13, or ACTv2018) seamlessly merged with 500 localized samples dynamically extracted from the target machine. To establish absolute baselines, we also tested the framework using the Target Local Data alone, TimeTrack alone, GWA-M13 alone, and ACTv2018 alone. 

\begin{figure*}
    \centering
    \includegraphics[width=1\linewidth]{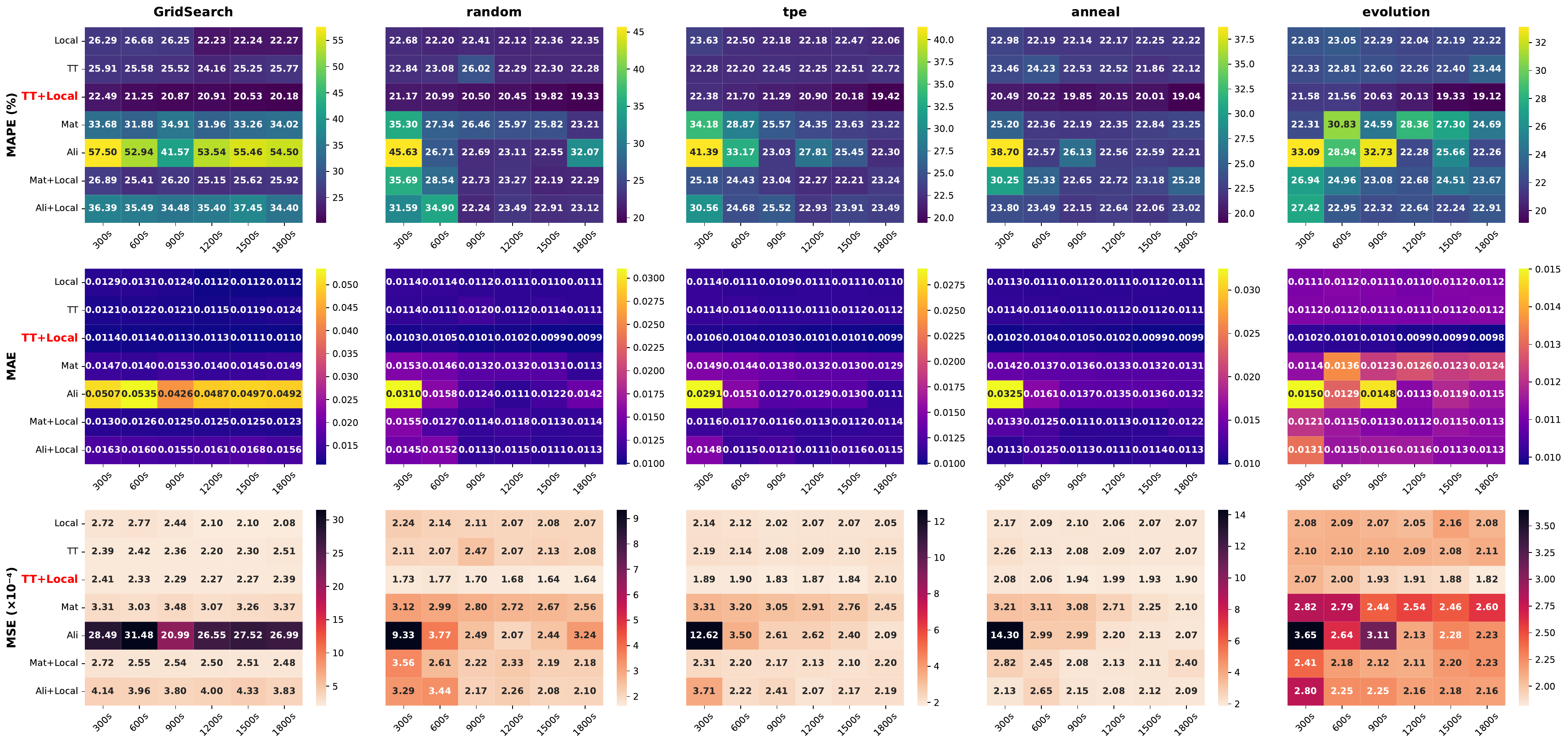}
    \caption{Performance heatmap of NAS CPU models comparing error metrics (MAPE, MAE, MSE) across varying search methods, durations, and data-mixing scenarios.}
    \label{fig:heatmap}
\end{figure*}

The empirical outcomes of these scenarios across the different time intervals are visualized in Figure \ref{fig:heatmap}. 

A granular analysis of the isolated baselines confirms that relying exclusively on coarse-grained public datasets yields severely degraded predictive accuracy. The Alibaba (\textit{ali\_only}) trace displays the worst overall baseline performance; at a 300-second search duration, its MAPE spans from an elevated 33.09\% under Evolutionary search to an appalling 57.49\% using Grid Search, accompanied by a high MSE of 0.002849. The Materna (\textit{mat\_only}) baseline behaves similarly, with Grid Search remaining stagnant at roughly 33.67\% to 34.01\% MAPE across all time intervals. We attribute this severe performance degradation to their coarse 5-minute sampling intervals, which mathematically act as a low-pass filter, smoothing out transient micro-bursts and high-frequency volatility. Consequently, models trained on these baselines lack the structural resolution to track sharp, short-term temporal variations. Conversely, the high-resolution TimeTrack baseline (\textit{tt\_only}) leverages its dense 45-second sampling interval to achieve significantly better baseline stability, stabilizing around 22.12\% to 22.28\% MAPE under TPE and Simulated Annealing. However, even this dense generic trace fails to outperform the Target Local Data alone (\textit{local\_only}), which hits a minimum MAPE of 22.04\% and an MSE of 0.000204 within 1200 seconds under Evolutionary search.

The most definitive and impactful finding of this evaluation is the universal efficacy of our proposed data-mixing framework. Across all test cases, merging target local data with any generic baseline yielded substantial performance gains over using those generic datasets in isolation. For instance, the \textit{local\_plus\_ali} and \textit{local\_plus\_mat} configurations successfully reclaimed predictive accuracy, dropping error rates well below their isolated counterparts. However, these configurations remained bounded by a distinct performance ceiling, failing to match the accuracy of the \textit{local\_plus\_tt} mixture. We attribute this performance delta directly to the telemetry collection intervals: while local data injects node-specific context, the underlying 5-minute sampling interval of the Alibaba and Materna traces fundamentally bottlenecks the model's ability to extract high-frequency structural motifs. Conversely, by blending local samples with the dense, 45-second intervals of TimeTrack, the \textit{local\_plus\_tt} configuration breaks the 20\% accuracy barrier. Under Simulated Annealing at 1800 seconds, it achieves the absolute best performance across the entire experiment, downscaling to a highly precise 19.04\% MAPE and a remarkably low MAE of 0.000994.

Interestingly, the data-mixing strategy also introduces an unexpected regularizing effect that alters search space topology. In the \textit{local\_plus\_tt} domain, Random Search matches sophisticated meta-heuristics, dropping smoothly from 21.17\% MAPE (300s) to 19.32\% MAPE (1800s), while yielding the lowest recorded variance with an MSE of 0.000163. This implies that high-resolution data-mixing smooths the loss landscape, making optimal architectural configurations highly abundant and accessible even to stochastic sampling routines.

Furthermore, evaluating performance across the three tracking metrics uncovers a vital diagnostic divergence between MAE and MSE. For instance, in the \textit{local\_plus\_mat} dataset at 1800 seconds, Simulated Annealing records a competitive MAE of 0.01219, yet its MSE deteriorates to 0.000240 compared to its 1500-second mark (0.000210). This indicates that while the model's average absolute error remained stable, it began generating isolated, high-magnitude prediction errors—a behavior that would trigger catastrophic SLA violations in a production environment. 

Finally, the temporal trajectories of the optimization strategies highlight a stark operational contrast between dynamic search heuristics and deterministic Grid Search. Across all datasets, heuristic and probabilistic strategies (TPE, Evolution, Annealing) demonstrate consistent, smooth asymptotic convergence as search duration increases from 300 to 1800 seconds. Grid Search, by contrast, suffers from severe dimensional paralysis. In the \textit{local\_only} scenario, Grid Search exhibits a sudden, step-function phase change, stagnating at $\sim$26.68\% MAPE before abruptly plummeting to 22.23\% at the 1200-second mark once it finally evaluates a viable hyperparameter coordinate. However, when confronted with sparse or mixed environments like \textit{local\_plus\_ali}, Grid Search remains totally paralyzed, flatlining at an unusable $\sim$34.40\% to 37.45\% MAPE across the entire 30-minute window. Exhaustive grid approaches waste precious CPU cycles evaluating known poor architectural configurations sequentially, proving that intelligent, adaptive search strategies are strictly necessary for resource-constrained, rapid edge deployment.

In summary, the empirical evidence demonstrates that our data-mixing approach successfully bootstraps initial model accuracy under strict computational and temporal constraints. However, while these results confirm the structural viability of blending datasets, this baseline evaluation relied on a static configuration of exactly 500 target local samples. To determine whether this specific volume represents an optimal operational threshold, or if predictive accuracy can be further optimized with a smaller local footprint or a different mixing ratio, it is necessary to analyze the framework's sensitivity to data volume. Consequently, we next isolate the localized data size as our primary independent variable, systematically evaluating its direct impact on NAS convergence and model accuracy to uncover the ideal data-provisioning boundaries.

\subsubsection{Sensitivity Analysis of Local Data Volume}
\label{eval:data_size}

While the previous evaluation used a fixed amount of local data, this experiment systematically varies the Target Local Data volume from 10 to 500 samples. This analysis serves two main purposes: first, to identify the minimum number of local samples required to generate accurate models; and second, to evaluate how effectively high-resolution generic data can compensate for data scarcity at the target node. For this study, we focus exclusively on Mean Absolute Percentage Error (MAPE) as our primary metric, as its scale-independent nature allows for a uniform comparison across different data sizes and search strategies.

\begin{figure*}
    \centering
    \includegraphics[width=01\linewidth]{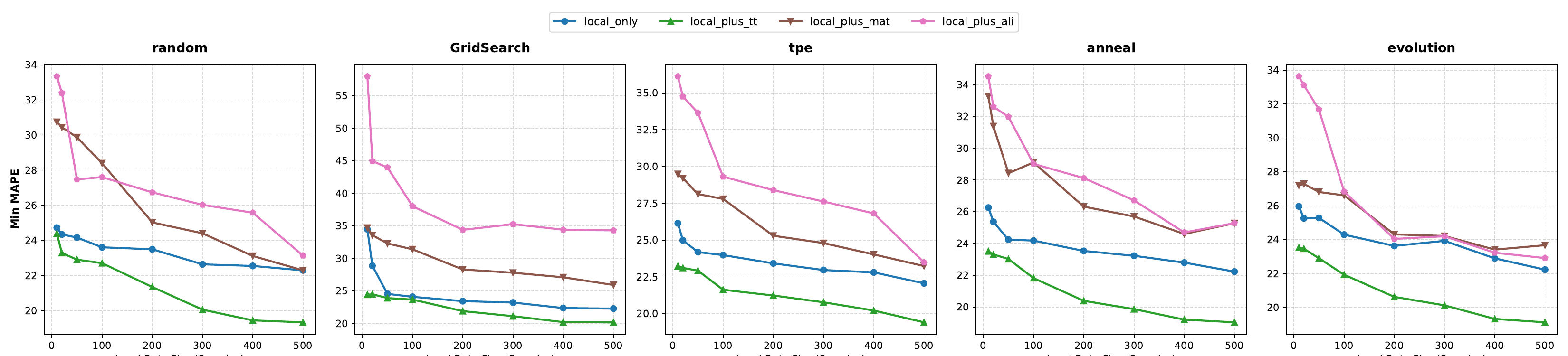}
    \caption {Impact of target data size (10 to 500 samples) on the minimum achieved MAPE across isolated and mixed data configurations using various NAS search strategies.}
    \label{fig:datasize}
\end{figure*}

The empirical trajectory of model accuracy as a function of localized data size is visualized in Figure \ref{fig:datasize}.

\begin{figure*}
    \centering
    \includegraphics[width=1\linewidth]{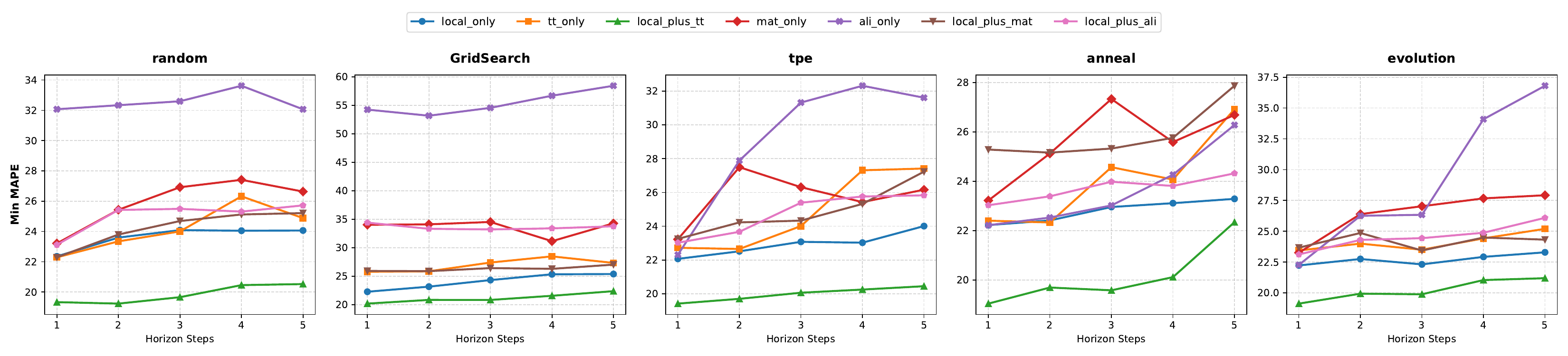}
    \caption{Minimum prediction error (MAPE) across different forecasting horizons (from 1 to 5 steps ahead) for all dataset combinations and search strategies.}
    \label{fig:horizon}
\end{figure*}

The resulting dataset exposes a clear, universal correlation: scaling the local data volume from 10 to 500 samples consistently drives down the minimum achieved MAPE across all data compositions. However, the performance delta between isolated and mixed data strategies at the ultra-sparse edge reveals the true structural value of the high-resolution baseline. In the \textit{local\_only} baseline configuration at 10 samples, the NAS engine struggles significantly, returning elevated MAPEs across all search regimes, with Grid Search recording a highly inaccurate 34.47\% MAPE and the best-performing stochastic search (Random) bounded at 24.72\% MAPE. This confirms that 10 telemetry points provide an entirely insufficient history for a network to deduce basic autoregressive properties in isolation.

This bottleneck is profoundly broken when introducing the \textit{local\_plus\_tt} configuration. By injecting 1,000 dense, high-resolution structural samples from TimeTrack, the pipeline achieves remarkable accuracy even when restricted to only 10 local samples. Under this extreme data constraint, TPE achieves a 23.23\% MAPE, and Random search marks 24.38\% MAPE. Strikingly, these mixed configurations utilizing a mere 10 local samples match or outperform the predictive accuracy of models trained on 500 samples of localized data alone (\textit{local\_only} at 500 samples hovers between 22.06\% and 22.28\% MAPE). This crossover effect proves that high-density generic data effectively shifts the search landscape, allowing the NAS framework to discover high-performing structural motifs that generalize perfectly to the target node despite near-zero initial local telemetry.

Conversely, analyzing the coarse-grained public mixtures highlights the strict limitations of large collection intervals. When mixed with 10 local samples, both \textit{local\_plus\_mat} and \textit{local\_plus\_ali} exhibit severe performance penalties. The Alibaba mixture (\textit{local\_plus\_ali}) shows a catastrophic 57.97\% MAPE under Grid Search and a restricted 36.12\% MAPE under TPE at the 10-sample mark. While scaling the local data volume to 500 samples does allow these coarse-grained mixtures to recover—eventually converging down to $\sim$22.29\% for Materna and $\sim$22.91\% for Alibaba—they never break the 20\% performance ceiling. In sharp contrast, the high-resolution TimeTrack mixture scales cleanly as local context accumulates, with Simulated Annealing unlocking the absolute global optimum of 19.04\% MAPE at the 500-sample milestone. This clear delta validates our hypothesis: while local data provides essential contextual alignment, the collection interval of the structural baseline dictates the model's ultimate accuracy ceiling. 

In summary, this sensitivity analysis demonstrates that mixing data creates an incredibly resilient pipeline capable of accurate deployments even during the earliest phases of target node bootstrapping. However, all evaluations conducted up to this point have operated under a single-step forecast horizon, testing only the immediate look-ahead capability of the models. While a 1-step horizon validates the fundamental data-mixing mechanics, real-world cluster management requires broader predictive horizons to safely absorb scheduling and migration overheads. Having thoroughly mapped the ideal data-provisioning boundaries, we proceed to our final experiment to evaluate how these data compositions withstand the compounding errors associated with multi-step-ahead forecasting.

\subsubsection{Predictive Accuracy Across Horizons}

In this experiment, we test how well our models predict multiple steps into the future. We vary the forecasting horizon from 1 step ahead up to 5 steps ahead. We run these tests across five different data setups: local data only (\texttt{local\_only}), external datasets alone (\texttt{tt\_only}, \texttt{mat\_only}, \texttt{ali\_only}), and mixed data combinations (\texttt{local\_plus\_tt}, \texttt{local\_plus\_mat}, \texttt{local\_plus\_ali}). For each setup, we also test various tuning methods like Random Search, Grid Search, TPE, Annealing, and Evolution to find the lowest possible error.

Predicting the immediate next step is usually easy for machine learning models because recent patterns are highly relevant. However, in real-world deployments, systems need to plan multiple steps in advance. This experiment is necessary because forecasting errors tend to compound and grow over longer time windows. By testing horizons 1 through 5, we can pinpoint exactly when a model's predictions stop being reliable and discover which data combinations best prevent the model from losing accuracy over time.

As shown in Figure~\ref{fig:horizon}, we evaluate the performance using the Mean Absolute Percentage Error (MAPE). The data reveals several critical trends:

\begin{itemize}
    \item \textbf{The Compounding Error Trend:} Across all datasets, error increases as the horizon grows from 1 to 5. For example, in the \texttt{local\_only} dataset, the best MAPE starts at $22.06\%$ (using TPE at horizon 1) and degrades to $23.28\%$ (using Annealing at horizon 5). This confirms the expected difficulty of longer-term forecasting.
    
    \item \textbf{The Clear Superiority of Telemetry Data:} The mixed dataset combining local and telemetry data (\texttt{local\_plus\_tt}) completely outperforms every other setup. At horizon 1, it achieves the lowest overall error in the entire experiment at $19.04\%$ (via Annealing). Remarkably, its error at horizon 5 ($20.45\%$) is still significantly lower than any other dataset can achieve even at horizon 1. This shows that telemetry data provides a strong, stable signal that protects the model from severe error growth.
    
    \item \textbf{Standalone Dataset Weaknesses:} Relying purely on external data without local context yields poor results over time. The \texttt{mat\_only} dataset starts at $23.21\%$ and jumps to $26.16\%$ by horizon 5. The \texttt{ali\_only} dataset degrades even worse, starting at $22.21\%$ and scaling up to a high error of $26.27\%$ at horizon 5. Grid Search on these standalone datasets performs exceptionally poorly, spiking up to $58.43\%$ MAPE on \texttt{ali\_only}.
    
    \item \textbf{Incompatible Combinations:} Interestingly, mixing local data with \texttt{mat} or \texttt{ali} data does not replicate the success of the telemetry mix. The \texttt{local\_plus\_mat} setup stays hovering around $22.29\%$ to $24.32\%$, showing that not all external data sources combine well with local node data.
\end{itemize}

The results of this horizon optimization test prove that multi-step forecasting accuracy depends heavily on the type of data used rather than just the optimization method. While all models suffer from rising errors as the prediction window lengthens, combining local data with telemetry data uniquely flattens this error growth curve. We can conclude that for multi-step forecasting tasks up to horizon 5, the \texttt{local\_plus\_tt} architecture is the only configuration that maintains a high level of predictive reliability for an initial model deployment.

\subsection{Post-Deployment Computational Footprint}
While high-resolution training drastically improves automated model generation, it inevitably forces the final deployed algorithm to process more data points in real-time. Specifically, a 45-second sampling interval yields $6.67\times$ more data over a given time window compared to a 5-minute interval. To ensure this does not violate edge computing constraints, we evaluated the real-time resource footprint of various fixed architectures (CNN, GRU, LSTM, RNN, Transformer) as data volume scaled up proportionally.

\begin{figure*}[htbp]
    \centering
    \includegraphics[width=\textwidth]{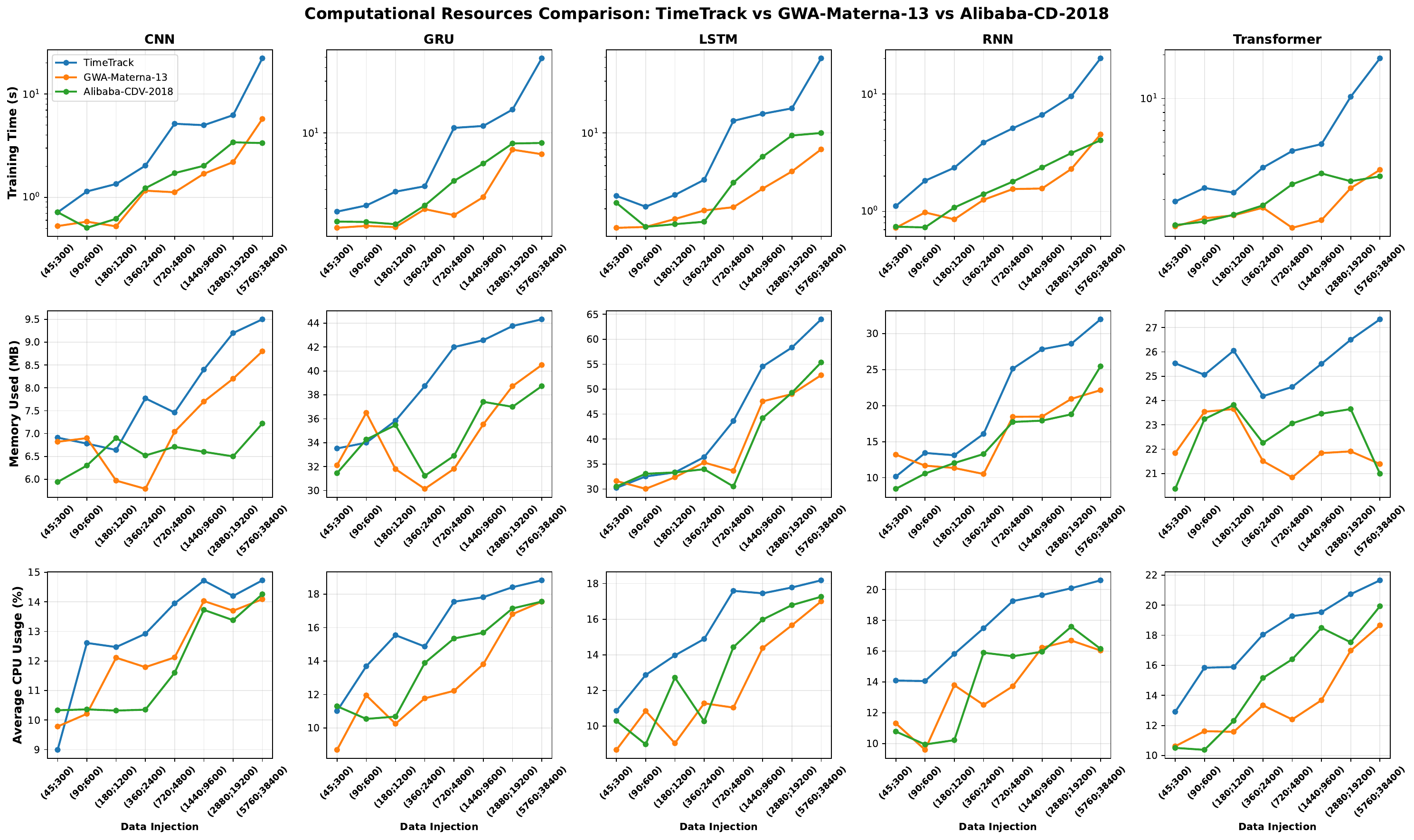}
    \caption{Training time, memory, and CPU usage for deployed models. The graphs scale proportionally to represent the $6.67\times$ data density difference of the 45-second TimeTrack interval vs. 5-minute intervals.}
    \label{fig:computation}
\end{figure*}

As depicted in Figure \ref{fig:computation}, processing the denser TimeTrack data logically increases computation time and memory usage compared to the sparse datasets. However, the absolute costs remain highly predictable and negligible for modern edge hardware. For instance, LSTMs—one of the most demanding architectures tested—required only \textbf{49 seconds} to process a massive window of 38,400 TimeTrack samples, peaking at roughly 50-55 MB of RAM. CPU usage remained similarly steady between 10-15\%. 

CNNs proved to be the most computationally efficient, while Transformers showed moderate memory usage that scaled predictably with sequence length. Ultimately, these results prove that the massive gains in prediction accuracy and NAS automation achieved by utilizing 45-second intervals far outweigh the minor, easily accommodated increase in post-deployment computational cost.

\section{Conclusion and Future Work}
\label{conclusion}
In this paper, we presented a fully automated predictive orchestration architecture designed to overcome the cold-start generalization problem and manage the extreme volatility of the Cloud-Edge Continuum. Our framework establishes a robust, data-driven foundation for proactive Zero Touch Management across highly heterogeneous environments.

To achieve this, we first introduced a lightweight, plugin-based Resource Exposer (RE) framework. The RE enables dynamic, infrastructure-agnostic telemetry extraction through a decentralized self-registration mechanism and a unified data format. Our performance evaluations demonstrated that the RE operates with minimal CPU and memory footprints and maintains low-latency API responses—even under aggressive 1-second data collection intervals—proving its exceptional suitability for resource-constrained edge and far-edge nodes. Furthermore, we solved the automated model generation challenge by merging this dynamic local telemetry with TimeTrack, our highly detailed, 45-second resolution structural dataset. We proved that leveraging TimeTrack as a foundational baseline within a Neural Architecture Search engine significantly improves both search efficiency and the predictive accuracy of the resulting forecasting models when compared to standard, coarse-grained public datasets. While processing high-resolution data inherently increases the post-deployment computational footprint, we demonstrated that this overhead remains highly predictable, stable, and well within acceptable bounds for modern edge hardware.

Despite these promising results, certain limitations remain that outline clear pathways for future work. 
\begin{enumerate}
    \item First, although our experiments demonstrated how the volume of localized target data injected into the pipeline directly influences the initial model accuracy, the current framework lacks a mechanism to dynamically calculate the optimal number of telemetry samples required to achieve a predefined target accuracy within a strict time constraint.
    \item  Second, the feature selection process is not yet automated; depending on the specific forecasting output and the telemetry metrics available via the Resource Exposer, the architecture cannot automatically determine which data columns should be included in the training phase based on cross-field correlations. 
\end{enumerate}

To transition toward a more automatic framework, our future research will focus on automating both dimensions: implementing an optimization layer to dynamically bound the required target sample sizes based on SLA performance targets, and developing a correlation-driven automated feature-selection mechanism to optimize input training configurations for maximum precision.

\bibliographystyle{unsrt}  


\bibliography{mybiblio}

@misc{openairinterface,
  author       = "{OpenAirInterface}",
  title        = "{OpenAirInterface: 5G Software Alliance for Democratising Wireless Innovation}",
  year         = 2023,
  url          = "https://openairinterface.org/",
  note         = "Accessed: 2024-10-30"
}

@misc{kagle,
  author       = {Abdelghani Meliani},
  title        = {TimeTrack: OpenAirInterface (OAI) CI/CD Time Series Dataset},
  year         = {2025},
  url          = {https://www.kaggle.com/datasets/abdelghanimeliani/open-air-interface-cicd-timeseries-dataset/data},
  note         = {Accessed: Feb 13, 2025}
}

@misc{zenodo,
  author       = {Abdelghani Meliani},
  title        = {TimeTrack: OpenAirInterface (OAI) CI/CD Time Series Dataset},
  year         = {2025},
  url          = {https://zenodo.org/records/15442251},
  note         = {Accessed: Aug 26, 2025}
}

@misc{openairinterface_test,
  author       = "{OpenAirInterface}",
  title        = "{OpenAirInterface Ci/Cd webpage}",
  year         = 2023,
  url          = "https://openairinterface.org/test-measurement/",
  note         = "Accessed: 2024-10-30"
}

@inproceedings{bagaa2026layer,
  title={Layer-Reuse Aware Optimization for Efficient Microservice Migration in UAV Edge Systems},
  author={Bagaa, Miloud and Ksentini, Adlen and others},
  booktitle={ICC 2026, IEEE International Conference on Communications},
  year={2026}
}

@conference{CECC,
  author = {Meliani, Abd Elghani and  Ksentini, Adlen and  Mekki, Mohamed and  Kadouma, Abdelhak and   Amaxilatis, Dimitrios and  Ba, Amadou and  Ojeda Coronado, Eduardo and  Beredimas, John and  Mouloss, Vrettos and  Sengupta, Souvik and  Klonidis, Dimitris and  Verikoukis, Christos},
  title = {AI-native CECC management architecture: Enabling scalable and intelligent cloud-edge computing},
  year = {2025},
  editor = {IEEE},
  address = {Athens},
  }

@ARTICLE{Shannon,
  author={Jerri, A.J.},
  journal={Proceedings of the IEEE}, 
  title={The Shannon sampling theorem—Its various extensions and applications: A tutorial review}, 
  year={1977},
  volume={65},
  number={11},
  pages={1565-1596},
  doi={10.1109/PROC.1977.10771}}

@misc{azure_public_dataset_v1,
  title = "{Microsoft Azure Public Dataset V1}",
  author = "{Microsoft Corporation}",
  year = 2019,
  url = "https://github.com/Azure/AzurePublicDataset/blob/master/",
  note = "Accessed: 2024-10-30"
}

@misc{azure_public_dataset_v2,
  title = "{Microsoft Azure Public Dataset V2}",
  author = "{Microsoft Corporation}",
  year = 2020,
  url = "https://github.com/Azure/AzurePublicDataset/blob/master/",
  note = "Accessed: 2024-10-30"
}

@misc{googleClusterData2011,
  author       = "{Google Inc.}",
  title        = "{Google Cluster Data 2019}",
  year         = 2019,
  url          = "https://github.com/google/cluster-data/blob/master",
  note         = "Accessed: 2024-08-27"
}

@misc{alibabaClusterTrace2018,
  author       = "{Alibaba Inc.}",
  title        = "{Alibaba Cluster Trace v2018}",
  year         = 2018,
  url          = "https://github.com/alibaba/clusterdata/blob/master/cluster-trace-v2018",
  note         = "Accessed: 2024-08-27"
}

@misc{gridWorkloadsArchive,
  author       = "{Delft University of Technology}",
  title        = "{Grid Workloads Archive GWA-T-13-Materna}",
  url          = "https://atlarge-research.com/gwa-t-13",
  note         = "Accessed: 2024-08-27"
}

@conference{timetrack,
  author = {Abd Elghani, Meliani and  Arora, Sagar and  Ksentini, Adlen and  Knopp, Raymond},
  title = {TimeTrack: A dataset for exploring temporal patterns and predictive insights into OpenAirInterface (OAI) CI/CD cluster},
  booktitle = {ICC 2025, IEEE International Conference on Communications 2025, 8-12 June 2025, Montreal, Canada},
  year = {2025},
  editor = {IEEE},
  address = {Montreal},
}

@conference{exposer,
  author = {Meliani, Abd Elghani and  Ksentini, Adlen},
  title = {Lightweight resource exposure framework for efficient service and resource orchestration in the cloud-edge continuum},
  booktitle = {ICC 2025, IEEE International Conference on Communications, 2nd Workshop on the Path Towards 6G: Standardization and Research Vision, 8-12 June 2025, Montreal, Canada},
  year = {2025},
  editor = {IEEE},
  address = {Montreal},
}

@article{migration,
title = {Resiliency focused proactive lifecycle management for stateful microservices in multi-cluster containerized environments},
journal = {Computer Communications},
volume = {236},
pages = {108111},
year = {2025},
issn = {0140-3664},
doi = {https://doi.org/10.1016/j.comcom.2025.108111},
url = {https://www.sciencedirect.com/science/article/pii/S0140366425000684},
author = {Abd Elghani Meliani and Mohamed Mekki and Adlen Ksentini},

}

@article{nas,
title = {A review of neural architecture search},
journal = {Neurocomputing},
volume = {474},
pages = {82-93},
year = {2022},
issn = {0925-2312},
doi = {https://doi.org/10.1016/j.neucom.2021.12.014},
author = {Dilyara Baymurzina and Eugene Golikov and Mikhail Burtsev},

}

@article{deeplearning,
  author    = {Yann LeCun and Yoshua Bengio and Geoffrey Hinton},
  title     = {Deep learning},
  journal   = {Nature},
  year      = {2015},
  volume    = {521},
  number    = {7553},
  pages     = {436--444},
  doi       = {10.1038/nature14539},
  url       = {https://doi.org/10.1038/nature14539},
  issn      = {1476-4687},
  month     = {May}
}

@article{lstm,
    author = {Hochreiter, Sepp and Schmidhuber, Jürgen},
    title = {Long Short-Term Memory},
    journal = {Neural Computation},
    volume = {9},
    number = {8},
    pages = {1735-1780},
    year = {1997},
    month = {11},
    issn = {0899-7667},
    doi = {10.1162/neco.1997.9.8.1735},
    url = {https://doi.org/10.1162/neco.1997.9.8.1735},
    eprint = {https://direct.mit.edu/neco/article-pdf/9/8/1735/813796/neco.1997.9.8.1735.pdf},
}

@phdthesis{libri2019towards,
  title={Towards High-Resolution Monitoring for HPC and Data Center Analytics, Automation and Control},
  author={Libri, Antonio},
  year={2019},
  school={ETH Zurich}
}

@inproceedings{chuah2006sampling,
  title={Is Sampled Data Sufficient for Anomaly Detection?},
  author={Mai, J. and others},
  booktitle={Proceedings of IMC},
  year={2006}
}

@article{garland2019effects,
  title={Effects of Resampled Data on Time Series Forecasting Accuracy},
  author={Garland, Jennifer},
  year={2019}
}

@software{nni2021,
   author = {{Microsoft}},
   month = {1},
   title = {{Neural Network Intelligence}},
   url = {https://github.com/microsoft/nni},
   version = {2.0},
   year = {2021}
}

@online{ietfmetrics,
    title = {Operational Compute Metrics},
    author = {Sabine Randriamasy and Luis M. Contreras and Jordi Ros-Giralt and Roland Schott},
    year = {2024-07-07},
    url = {https://datatracker.ietf.org/doc/draft-rcr-opsawg-operational-compute-metrics/},
    urldate = {2024-09-09},
}

@online{prometheus,
    title = {Prometheus: Monitoring System and Time Series Database},
    author = {{Prometheus}},
    year = {n.d.},
    url = {https://prometheus.io/},
    urldate = {2024-09-09},
}

@online{nagios,
    author = {{Nagios}},
    title = {Nagios - The Industry Standard in IT Infrastructure Monitoring},
    year = {2024},
    url = {https://www.nagios.org/},
    note = {Accessed: 2024-09-12}
}

@INPROCEEDINGS{decor,
  author={Shen, Shan-Hsiang and Akella, Aditya},
  booktitle={2012 IEEE 20th International Workshop on Quality of Service}, 
  title={DECOR: A distributed coordinated resource monitoring system}, 
  year={2012},
  volume={},
  number={},
  pages={1-9},
  doi={10.1109/IWQoS.2012.6245994}}

@ARTICLE{mekki,
  author={Mekki, Mohamed and Arora, Sagar and Ksentini, Adlen},
  journal={IEEE Transactions on Network and Service Management}, 
  title={A Scalable Monitoring Framework for Network Slicing in 5G and Beyond Mobile Networks}, 
  year={2022},
  volume={19},
  number={1},
  pages={413-423},
  doi={10.1109/TNSM.2021.3119433}}

@article{dprof,
  title={Dprof-Distributed System Profiling and Tracing},
  author={Nguyen, Tuan and Pandey, Manish}
}

@incollection{alla2020mlops,
  title={What is mlops?},
  author={Alla, Sridhar and Adari, Suman Kalyan},
  booktitle={Beginning MLOps with MLFlow: Deploy Models in AWS SageMaker, Google Cloud, and Microsoft Azure},
  pages={79--124},
  year={2020},
  publisher={Springer}
}

@article{karmaker2021automl,
  title={Automl to date and beyond: Challenges and opportunities},
  author={Karmaker, Shubhra Kanti and Hassan, Md Mahadi and Smith, Micah J and Xu, Lei and Zhai, Chengxiang and Veeramachaneni, Kalyan},
  journal={Acm computing surveys (csur)},
  volume={54},
  number={8},
  pages={1--36},
  year={2021},
  publisher={ACM New York, NY}
}



\end{document}